\documentclass[11pt,a4paper]{article}
\usepackage[hyperref]{acl2020}
\usepackage{times}
\usepackage{latexsym}

\usepackage{amsfonts}
\usepackage{amsmath}
\usepackage{amssymb}
\newcommand\DNAME{ALICE}

\usepackage{microtype}

\aclfinalcopy 
\def\aclpaperid{22} 

\title{Adversarial Training for Commonsense Inference}
\author{
Lis Pereira\textsuperscript{1}, Xiaodong Liu\textsuperscript{2}, Fei Cheng\textsuperscript{3}, Masayuki Asahara\textsuperscript{4}, Ichiro Kobayashi\textsuperscript{1}
 \\ 
  \textsuperscript{1} Ochanomizu University
  \textsuperscript{2} Microsoft Research 
    \textsuperscript{3} Kyoto University \\
    \textsuperscript{4} The National Institute for Japanese Language and Linguistics (NINJAL)
 \\
  kanashiro.pereira@ocha.ac.jp,  xiaodl@microsoft.com, feicheng@i.kyoto-u.ac.jp\\ masayu-a@ninjal.ac.jp,  koba@is.ocha.ac.jp 
}
\date{}

\begin{document}
\maketitle
\begin{abstract}
We propose an \textbf{A}dversaria\textbf{L} training algorithm for commonsense \textbf{I}nferen\textbf{CE} ({\DNAME}).
We apply small perturbations to word embeddings and minimize the resultant adversarial risk to regularize the model. We exploit a novel combination of two different approaches to estimate these perturbations: 1) using the true label and 2) using the model prediction. Without relying on any human-crafted features, knowledge bases or additional datasets other than the target datasets, our model boosts the fine-tuning performance of RoBERTa, achieving competitive results on multiple reading comprehension datasets that require commonsense inference. 

\end{abstract}
\section{Introduction}

Commonsense knowledge is often necessary for natural language understanding. As shown in Table \ref{example}, we can understand
that the writer needs help to get dressed and seems upset with this situation, indicating that he or she is probably not a child. Thus, we can infer that a possible reason that the writer needs to be dressed by other people is that he or she may have a physical disability \cite{huang2019cosmos}. Although a simple task for humans, it is still challenging for computers to understand and reason about commonsense.

\begin{table}[h]
\begin{center}
\begin{tabular}{|p{72.5mm}|}
\hline \textit{Paragraph}: It's a very humbling experience when you need someone to dress you every morning, tie your shoes, and put your hair up. Every menial task takes an unprecedented amount of effort. It made me appreciate Dan even more. But anyway I shan't dwell on this (I'm not dying after all) and not let it detract from my lovely 5 days with my friends visiting from Jersey.  \\ \\ 

\textit{Question}: What's a possible reason the writer needed someone to dress him every morning? \\ \\

\textit{Option1}: The writer doesn't like putting effort into these tasks. \\

\textit{Option2}: \bf{The writer has a physical disability.} \\

\textit{Option3}: The writer is bad at doing his own hair. \\

\textit{Option4}: None of the above choices. \\
\hline
\end{tabular}
\end{center}
\caption{\label{example} Example from the CosmosQA dataset \cite{huang2019cosmos}. The task is to identify the correct answer option. The correct answer is in \bf{bold}.}
\vspace{-5mm}
\end{table}

Commonsense inference in natural language processing (NLP) is generally evaluated via machine reading comprehension task, in the format of selecting plausible responses with respect to natural language queries. Recent approaches are based on the use of pre-trained Transformer-based language models such as BERT \cite{devlin2018bert}. Some approaches rely solely on these models by adopting either a single or multi-stage fine-tuning approach (by fine-tuning using additional datasets in a step-wise manner) ~\cite{li2019,sharma2019iit,liu2019blcu,huang2019cosmos,zhou2019going}, while others further enhance their word representations with knowledge bases such as ConceptNet~\cite{jain2019karna,da2019jeff,wang2020kadapter}. However, due to the often limited data from the downstream tasks and the extremely high complexity of the pre-trained model, aggressive fine-tuning can easily make the adapted model overfit the data of the target task, making it unable to generalize well on unseen data \cite{jiang2019smart}. Moreover, some researchers have shown that such pre-trained models are vulnerable to adversarial attacks \cite{jin2019bert}. 

Inspired by the recent success of adversarial training in NLP ~\cite{zhu2019freelb,jiang2019smart}, our \textbf{A}dversaria\textbf{L} training algorithm for commonsense \textbf{I}nferen\textbf{CE} ({\DNAME}) focuses on improving the generalization of pre-trained language models on downstream tasks by enhancing their robustness in the embedding space. More specifically, during the fine-tuning stage of Transformer-based models, e.g. RoBERTa \cite{liu2019roberta}, random perturbations are added to the embedding layer to regularize the model by updating the parameters on these adversarial embeddings. ALICE exploits a novel way of combining two different approaches to estimate these perturbations: 1) using the true label and 2) using the model prediction. Experiments show that we were able to boost the performance of RoBERTa on multiple reading comprehension datasets that require commonsense inference, achieving competitive results with state-of-the-art approaches.

\section{ALICE}
\label{method}
Given a dataset $D$ of $N$ training examples, $D=\{(x_1, y_1), (x_2, y_2), ..., (x_N, y_N)\}$, the objective of supervised learning is to learn a function $f(x; \theta)$ that minimizes the empirical risk, which is defined by $\min_{\theta} \mathbb{E}_{(x, y)\sim D}[l(f(x; \theta), y)]$. Here, the function $f(x; \theta)$ maps input sentences $x$ to an output space $y$, and $\theta$ are learnable parameters. While this objective is effective to train a neural network, it usually suffers from overfitting and poor generalization to unseen cases \cite{goodfellow2014explaining,madry2017towards}. To alleviate these issues, one can use adversarial training, which has been primarily explored in computer vision \cite{goodfellow2014explaining,madry2017towards}. The idea is to perturb the data distribution in the embedding space by performing adversarial attacks. Specifically, its objective is defined by:
\begin{equation}
\min_{\theta} \mathbb{E}_{(x, y)\sim D}[\max_{\delta} l(f(x + \delta; \theta), y)], 
\label{eq:advt}
\end{equation}
where $\delta$ is the perturbation added to the embeddings. One challenge of adversarial training is how to estimate this perturbation $\delta$, which is to solve the inner maximization, $\max_{\delta}l(f(x+\delta;\theta),y)$. A feasible solution is to approximate it by a fixed number of steps of a gradient-based optimization approach \cite{madry2017towards}.

Based on recent successful cases that applied adversarial training to NLP \cite{jiang2019smart,miyato2018virtual}, the approaches to estimate $\delta$ can be divided into two categories: adversarial training that uses the label $y$ \cite{zhu2019freelb} and adversarial training that uses the model prediction $f(x; \theta)$, i.e. a "virtual" label \cite{miyato2018virtual,jiang2019smart}. We hypothesize that these two categories complement each other: 
the first one is to improve the robustness of our target label, by avoiding an increase in the error of the unperturbed inputs, while the second term enforces the smoothness of the model, encouraging the output of the model not to change much, when injecting a small perturbation to the input. Thus, {\DNAME} proposes a novel algorithm by combining these two approaches, which is defined by:
\vspace{-2mm}
\begin{equation}
\begin{split}
\min_{\theta} \mathbb{E}_{(x, y)\sim D}[\max_{
\delta_1} l(f(x + \delta_1; \theta), y) + \\ \alpha \max_{
\delta_2} l(f(x + \delta_2; \theta), f(x; \theta))], 
\label{eq:alice}
\end{split}
\vspace{-2mm}
\end{equation}
where $\delta_1$ and $\delta_2$ are two perturbations, bounded by a general $l_p$ norm ball, estimated by a fixed $K$ steps of the gradient-based optimization approach. In our experiments, we set $p=\infty$. It has been shown that a larger $K$ can lead to a better estimation of $\delta$ \cite{qin2019adversarial,madry2017towards}. However, this can be expensive, especially in large models, e.g. BERT and RoBERTa. Thus, $K$ is set to 1 for a better trade-off between speed and performance. Note that $\alpha$ is a hyperparameter balancing these two loss terms. In our experiments, we set $\alpha$ to 1.
\section{Experiments}
\label{sec:exp}
\vspace{-2mm}
 
 \begin{table*}[h]
\begin{center}
\begin{tabular}{@{\hskip2pt}l@{\hskip1pt}|c|c|c|c|c|@{\hskip1pt}c@{\hskip2pt}}
\hline \bf Dataset & \bf \#Train & \bf \#Dev & \bf \#Test&\bf\#Label&\bf Task&\bf Metrics\\ \hline
CosmosQA & 25,262 & 2,985  & 6,963 &4&Relevance Ranking&Accuracy\\
MCScript2.0 & 14,191 & 2,020  & 3,610&2&Relevance Ranking &Accuracy\\
MCTACO & - & 3,783& 9,442&2&Pairwise Text Classification&Exact Match (EM)/F1\\

\hline
\end{tabular}
\end{center}
\caption{\label{dataset-table} Summary of the three datasets: CosmosQA, MCScript2.0 and MCTACO.}
\vspace{-4mm}
\end{table*}

\subsection{Datasets and Evaluation Metrics}
\vspace{-2mm}
We evaluate {\DNAME} on three reading comprehension benchmarks that require commonsense inference: 

\noindent{\bf CosmosQA} \cite{huang2019cosmos}:  a large-scale dataset that focuses on people's everyday narratives, asking questions about the likely causes or effects of events that require reasoning beyond the exact text spans in the context. It has 35,888 questions on 21,886 distinct contexts taken from blogs of personal narratives. Each question has four answer candidates, one of which is correct. 93.8\% of the dataset requires contextual commonsense reasoning. 

\noindent{\bf MCScript2.0} \cite{ostermann2019commonsense}: a dataset focused on short narrations on different everyday activities (e.g. baking a cake, taking a bus, etc.). It has 19,821 questions on 3,487 texts. Each question has two answer candidates, one of which is correct. Roughly half of the questions require inferences over commonsense knowledge. 

\noindent{\bf MC-TACO} \cite{zhou2019going}: a dataset that entirely focuses on a specific reasoning capablity: temporal commonsense. It considers five temporal properties, (1) duration (how long an event takes), (2) temporal ordering (typical order of events), (3) typical time (when an event occurs), (4) frequency (how often an event occurs), and (5) stationarity (whether a state is maintained for a very long time or indefinitely). It contains 13k tuples, each consisting of a sentence, a question, and a candidate answer, that should be judged as plausible or not. The sentences are taken from different sources such as news, Wikipedia and textbooks.

The summary of the datasets is in Table \ref{dataset-table}. For the MCTACO dataset, no training set is available. Following \citep{zhou2019going}, we use the dev set for fine-tuning the model. We perform 5-fold cross-validation for fine-tuning the parameters.

We evaluate CosmosQA and MCScript2.0 in terms of accuracy. Following \citep{ostermann2019mcscript2}, we also report for the MCScript2.0 accuracy on the commonsense based questions and accuracy on the questions that are not commonsense based. For the MCTACO, we report the exact match (EM) and F1 scores, following \citep{zhou2019going}. EM measures how many questions a system correctly labeled all candidate answers, while F1 measures the average overlap between one’s predictions and the ground truth.
Our implementation for pairwise text classification and relevance ranking tasks are based on the MT-DNN framework\footnote{https://github.com/namisan/mt-dnn} \cite{liu2019multi,liu2020mtmtdnn}. 

\subsection{Implementation Details}
\vspace{-1mm}
The RoBERTa\textsubscript{LARGE} model \cite{liu2019roberta} was used as the text encoder. We used ADAM \cite{kingma2014adam} as our optimizer with a learning rate in the range $\in \{1 \times 10^{-5}, 2 \times 10^{-5}, 3 \times 10^{-5}, 5 \times 10^{-5}, 5 \times 10^{-5}\}$ and a batch size $\in \{16, 32, 64\}$. The maximum number of epochs was set to 10. A linear learning rate decay schedule with warm-up over 0.1 was used, unless stated otherwise. We also set the dropout rate of all the task specific layers as 0.1, except 0.3 for MCTACO. To avoid gradient exploding, we clipped the gradient norm within 1. All the texts were tokenized using wordpieces and were chopped to spans no longer than 512 tokens. 





\subsection{Baselines}
\begin{table*}[h]
\begin{center}
\begin{tabular}{p{65mm}|c |c c c |c c}
\hline \bf  & \bf CosmosQA &   \multicolumn{3}{c}{\textbf{MCScript2.0}} & \multicolumn{2}{|c}{\textbf{MCTACO}}\\ \hline 
 Model & Acc & Acc &Acc\textsubscript{cs} &Acc\textsubscript{ood}  &EM&F1\\ \hline
\hline\multicolumn{7}{c}{\textbf{Development Set Results}} \\ \hline \hline
RoBERTa\textsubscript{LARGE} \cite{liu2019roberta} & 80.60  & 90.0&87.7&92.1 &44.12 &64.85\\
ADV  &81.14&92.1&89.8&94.3& 52.70& 78.12  \\ 
SMART \cite{jiang2019smart} &82.00&93.6& \textbf{91.8} &95.2 &53.79& 78.31 \\
{\DNAME} &\textbf{83.60}& \textbf{93.8}& 91.7 & \textbf{95.7}&\textbf{58.02}&\textbf{78.64}\\
\hline\multicolumn{7}{c}{\textbf{Test Set Results}} \\ \hline \hline 
 Human  Performance&94.00& 97.0&-&-&75.80&87.10\\ \hline 
\small{BERT + unit normalization} \cite{zhou2019going} & - & - &-&-& 42.70&69.90\\
T5-3B fine-tuned + number normalization\textsuperscript{*} & - & - &-&-& \textbf{59.08}&79.46\\
PSH-SJTU \cite{li2019} &-& 90.6&90.3&91.5&-&-\\
K-ADAPTER \cite{wang2020kadapter} &81.83& -&-&-&-&-\\  
GB-KSI(v2)\textsuperscript{*}&83.97& -&-&-&-&-\\  \hline
RoBERTa\textsubscript{LARGE} \cite{liu2019roberta}&  -&88.8 & 87.0& 90.7& 51.05& 76.85\\
ADV  &-&91.1&90.3&92.0 &54.27 &77.23  \\
SMART \cite{jiang2019smart} &81.90&91.8& 90.5 & 93.4&54.80& 78.03 \\ 
{\DNAME} &\textbf{84.57}&\textbf{92.5}& \textbf{91.6}&\textbf{93.5} &56.45&\textbf{79.50}\\ \hline
\hline
\end{tabular}
\end{center}
\vspace{-2mm}
\caption{\label{tab:dev}  Development and test results of CosmosQA, MCScript 2.0 and MCTACO. The best results are in \textbf{bold}. Note that RoBERTa\textsubscript{LARGE}, SMART and ALICE models use RoBERTa\textsubscript{LARGE} as the text encoder, and for a fair comparison, all these results are produced by ourselves.
On the test results, note that CosmosQA and MCTACO are scored by using the official evaluation server (https://leaderboard.allenai.org/). * denotes unpublished work and scores were obtained from the evaluation server on April 16, 2020. Acc\textsubscript{cs} denotes the accuracy on commonsense based questions and Acc\textsubscript{ood} denotes the accuracy on questions that are not commonsense based, i.e. out-of-domain questions.}
\vspace{-4mm}
\end{table*}



We compare ALICE to a list of state-of-the-art models, as shown in Table \ref{tab:dev}. \textbf{BERT + unit normalization} \cite{zhou2019going} is the BERT base model. The authors further add unit normalization to temporal expressions in candidate answers and fine-tune on the MC-TACO dataset. \textbf{RoBERTa\textsubscript{LARGE}} is our re-implementation of the large RoBERTa model by \cite{liu2019roberta}. \textbf{PSH-SJTU} \cite{li2019} is based on multi-stage fine-tuning XLNET \cite{yang2019xlnet} on RACE \cite{lai2017race}, SWAG \cite{zellers2018swag} and MC-Script2.0 datasets. \textbf{K-ADAPTER} \cite{wang2020kadapter} further enhances RoBERTa word representations with multiple knowledge sources, such as factual knowledge obtained through Wikipedia and Wikidata and linguistic knowledge obtained through dependency parsing web texts. \textbf{SMART} \cite{jiang2019smart} is an adversarial training model for fine-tuning pre-trained language models through regularization. SMART uses the model prediction, $f(x; \theta)$, for estimating the perturbation $\delta$. This model recently obtained state-of-the-art results on a bunch of NLP tasks on the GLUE benchmark \cite{wang2018glue}. We also compare ALICE with a baseline that uses only the label $y$ for estimating the perturbation $\delta$ (called model \textbf{ADV} hereafter) \cite{madry2017towards}.
\subsection{Results}
The results are summarized in Table \ref{tab:dev}. Overall, we observed that adversarial methods, i.e. ADV, SMART and ALICE, were able to achieve competitive results over the baselines, without using any additional knowledge source, and without using any additional dataset other than the target task datasets. These results suggest that adversarial training lead to a more robust model and help generalize better on unseen data. 

ALICE consistently oupterformed SMART (which overall outperformed ADV) across all three datasets on both dev and test sets, indicating that adversarial training that uses the label $y$ and adversarial training that uses the model prediction $f(x; \theta)$ are complementary, leading to better results. For example, on the CosmosQA dataset, we obtained a dev-set accuracy of 83.6\% with ALICE, a 1.6\% and 3.0\% absolute gains over SMART and RoBERTa\textsubscript{LARGE}, respectively. On the blind test-set, ALICE outperforms by a large margin K-ADAPTER, a model that enhances RoBERTa word representations with multiple knowledge sources. Our submission to the CosmosQA leaderboard achieved a test-set accuracy of 84.57\%, ranking first place among all submissions (as of April 16, 2020). On the MCScript2.0 dataset, ALICE obtained a dev-set accuracy of 93.8\% in total, a 0.2\% and 3.8\% absolute gains over SMART and RoBERTa\textsubscript{LARGE}, respectively. On the commonsense based questions, ALICE underperformed SMART by 0.1\% and outperformed RoBERTa\textsubscript{LARGE} by 4.0\%. On the out-of-domain questions, ALICE obtained 0.5\% and 3.6\% absolute gains over SMART and RoBERTa\textsubscript{LARGE}, respectively. On the MCScript2.0 test-set, ALICE outperformed all baselines (on all types of questions), including SMART, indicating that it can generalize better to unseen cases. Moreover, ALICE outperformed PSH-SJTU, which is based on XLNET, and used additional datasets other than MCScript2.0, while ALICE does not use any additional dataset. On the MCTACO dataset, ALICE obtained on the dev-set 56.20\% EM score, a 2.41\% and 12.8\% absolute gains over SMART and RoBERTa\textsubscript{LARGE}, respectively, and 79.06\% F1 score, a 0.75\% and 14.21\% absolute gains over SMART and RoBERTa\textsubscript{LARGE}, respectively. On the test-set, ALICE outperformed SMART obtaining absolute gains of 1.65\% and 1.47\% on EM and F1 scores, respectively. Compared to the T5-3B fine-tuned + number normalization model, which uses T5, a much larger model (with 3B parameters) than RoBERTa (300M parameters), ALICE obtained competitive results, outperforming by 0.04 on F1 score and obtaining 2.63\% lower score on EM.
Regarding the training time, ALICE takes on average 4X more time to train compared to standard fine-tuning. 


\section{Conclusion}
\vspace{-2mm}
We proposed {\DNAME}, a simple and efficient adversarial training algorithm for fine-tuning large scale pre-trained language models. Our experiments demonstrated that it achieves competitive results on multiple machine reading comprehension datasets, without relying on any additional resource other than the target task dataset. Although in this paper we focused on the machine reading comprehension task, {\DNAME} can be generalized to solve other downstream tasks as well, and we will explore this direction as future work.

\section*{Acknowledgments}
We thank the reviewers for their helpful feedback. This work has been supported by the project KAKENHI ID: 18H05521. \\

\bibliography{acl2020}
\bibliographystyle{acl_natbib}
\end{document}